\documentclass[letterpaper, 10 pt, conference]{ieeeconf}  

\IEEEoverridecommandlockouts                              

\usepackage[bookmarks=true,pagebackref=true,colorlinks=true]{hyperref}
\usepackage{amsmath,amssymb,array,balance,booktabs,cite,changes,color,enumerate,float,graphicx,hyperref,multicol,multirow,siunitx,times,subcaption,mathtools,colortbl}
\usepackage{wrapfig}
\usepackage[export]{adjustbox}
\usepackage{xcolor}
\usepackage[font=footnotesize]{caption}
\usepackage[export]{adjustbox}
\usepackage{algorithm}
\usepackage[noend]{algpseudocode}
\usepackage{xspace}
\definecolor{myblue}{rgb}{0.44, 0.65, 0.82}
\definecolor{mygreen}{rgb}{0.032, 0.6392, 0.2039}
\newcommand{\videourl}{\url{https://sites.google.com/berkeley.edu/aprl}}

\newcommand{\alert}[1]{\textbf{}}

\newcommand{\BEAS}{\begin{eqnarray*}}
\newcommand{\EEAS}{\end{eqnarray*}}
\newcommand{\BEA}{\begin{eqnarray}}
\newcommand{\EEA}{\end{eqnarray}}
\newcommand{\BEQ}{\begin{equation}}
\newcommand{\EEQ}{\end{equation}}
\newcommand{\BIT}{\begin{itemize}}
\newcommand{\EIT}{\end{itemize}}
\newcommand{\BNUM}{\begin{enumerate}}
\newcommand{\ENUM}{\end{enumerate}}
\newcommand{\BEL}[1]{\begin{equation}\label{#1}}
\newcommand{\EEL}{\end{equation}}

\newcommand{\state}{\mathbf{s}}

\newcommand{\action}{\mathbf{a}}

\newcommand{\statespace}{\mathcal{S}}
\newcommand{\actionspace}{\mathcal{A}}

\newcommand{\policy}{\pi}

\newcommand{\BA}{\begin{array}}
\newcommand{\EA}{\end{array}}

\DeclareMathOperator*{\expec}{\mathbb{E}}

\hypersetup{colorlinks = true, breaklinks = true, citecolor = [rgb]{0,0.07843,0.45098}, urlcolor = [rgb]{0,0.07843,0.45098}, linkcolor = [rgb]{0,0.07843,0.45098}} 

\newcommand{\metabbr}{APRL\xspace}

\title{\LARGE \bf Grow Your Limits: Continuous Improvement with \\Real-World RL for Robotic Locomotion}

\author{
\authorblockN{Laura Smith\textsuperscript{* 1}, Yunhao Cao\textsuperscript{* 1}, Sergey Levine\textsuperscript{1}}
\authorblockA{\textsuperscript{*}Equal contribution \textsuperscript{1}Berkeley AI Research, UC Berkeley\\
\texttt{\{smithlaura, caoyh2001\}@berkeley.edu}}
}

\begin{document}
\setlength{\textfloatsep}{7pt}

\maketitle
\thispagestyle{empty}
\pagestyle{empty}

\begin{abstract}
Deep reinforcement learning (RL) can enable robots to autonomously acquire complex behaviors, such as legged locomotion. However, RL in the real world is complicated by constraints on efficiency, safety, and overall training stability, which limits its practical applicability. We present~\metabbr, a policy regularization framework that modulates the robot's exploration over the course of training, striking a balance between flexible improvement potential and focused, efficient exploration. 
\metabbr enables a quadrupedal robot to efficiently learn to walk entirely in the real world within minutes and continue to improve with more training where prior work saturates in performance. We demonstrate that continued training with \metabbr results in a policy that is substantially more capable of navigating challenging situations and is able to adapt to changes in dynamics with continued training. Videos and code to reproduce our results are available at: \videourl
\end{abstract}

\section{Introduction}
\label{sec:intro}

The real world is noisy, diverse, unpredictable and challenging. To confront this complexity, a robot must be capable of versatile and robust behaviors. Designing controllers that anticipate and handle \textit{any} scenario a robot will encounter in its lifetime, whether through manual engineering or learning, is impractical. An alternative to providing a robot with predetermined capabilities is to instead equip it with the ability to autonomously improve during deployment, learning from its mistakes as it encounters unforeseen circumstances. Legged robots in particular have incredible mobility and can reach places challenging or unsuitable for humans, for example, in search and rescue scenarios, and thus must be robust to these unforeseen situations. 

Reinforcement learning (RL) offers a general framework for such a `self-improving robot,' providing a data-driven approach for learning behaviors through interaction. However, the practical application of end-to-end RL in robotic systems is often not straightforward~\cite{Zhu2020TheIO, Ibarz2021HowTT},
with many of the challenges stemming from high sample complexity: RL algorithms are notoriously data-hungry, while real-world data collection is notoriously expensive due to human supervision requirements, hardware damage, and other physical constraints. Although recent works~\cite{Kohl2004PolicyGR, Finn2017DeepVF, Kalashnikov2018QTOptSD, Levine2018LearningHC, Kalashnikov2021MTOptCM, Ebert2022BridgeDB, Zhang2019SOLARDS, Nagabandi2019DeepDM} have demonstrated encouraging progress toward end-to-end RL on real-world systems, often by applying the latest advances in sample-efficient RL, the efficiency and final performance of these methods still presents difficulties for persistent and reliable deployment on real-world platforms such as legged robots. In this work, we consider the task of learning quadrupedal locomotion and ask: how can we enable a robot to learn more agile locomotion in the real world, where it must learn and adapt efficiently amid diverse, challenging scenarios?

\begin{figure}[t]
    \centering
    \includegraphics[width=\linewidth]{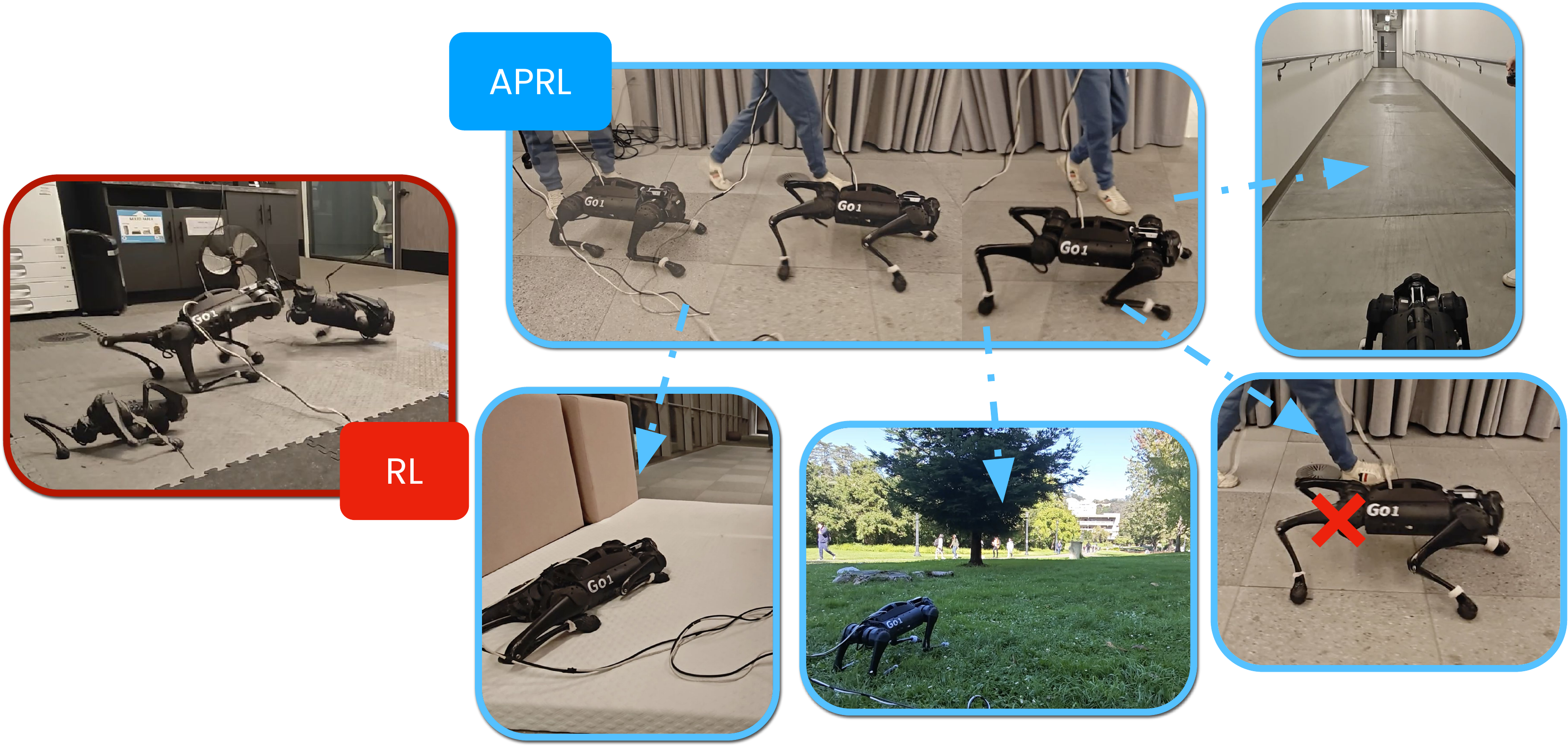}
    \caption{\footnotesize \metabbr uses a novel action space regularization technique based on dynamics prediction error to modulate exploration over the course of training. This enables real-world quadrupedal learning that can traverse challenging terrains and continually adapt to changes in dynamics. 
    }
    \label{fig:teaser}
    \vspace{-.2cm}
\end{figure}

We present~\metabbr, a system that addresses the real-world challenges of \textit{efficiency} and \textit{continual improvement} in robot learning via adaptive policy regularization, with a focus on quadrupedal locomotion. Our key observation is that the policy's search space of actions has a significant effect on the robot's ability to learn effectively. To illustrate this, consider tasking the robot to learn to walk with no prior knowledge. With full command of its joint range, most random behaviors will lead to catastrophic failure, and exploration becomes practically infeasible (see the behavior pictured on the left in~\autoref{fig:teaser}), but manually defining an appropriate space of actions to explore restricts the robot's capabilities, as the highest performance later in training might necessitate exceeding such restrictions. Our approach is to \textit{dynamically regularize} the policy, providing it with enough freedom to explore and improve, but not so much that its exploration is needlessly expensive. At the start of training, \metabbr biases the policy toward small-magnitude actions to avoid the robot having to first learn this through costly first-hand experience, but as the robot becomes more competent, it should be allowed to explore more aggressive actions. To this end, we introduce an adaptive penalty based on how \emph{familiar} the current situation is for the robot. This adaptive strategy allows the robot to explore more aggressively once it has learned about its surroundings, and dial back its exploration it it encounters unexpected dynamics, where we would expect the policy to not generalize well. We measure familiarity by training a dynamics model on the data the robot has collected and using its prediction error to dictate the policy's recommended search space. We use this action regularization synergistically with resetting the agent~\cite{Nikishin2022ThePB}, which combats the early overfitting, a common problem sample-efficient RL algorithms are prone to. Resets improve plasticity, providing more opportunities to learn when needed, and the dynamic penalty on actions focuses the robot's behavior, i.e., to prevent excessive falling and inefficient exploration.

In summary, the main contribution of our work is a system,~\metabbr, for efficiently learning quadrupedal locomotion directly in the real world using a novel, adaptive regularization strategy that promotes more efficient, high-performing, and stable training. We demonstrate that on a 12 degree-of-freedom Unitree Go1 quadruped, our system can learn to walk forward in just minutes starting completely from scratch. While prior work saturates in performance, \metabbr allows the robot to continuously improve with more training. Furthermore, the behavior learned with \metabbr performs significantly better (on average by a factor of 2) in terms of its average walking speed in challenging situations as shown in~\autoref{fig:teaser}, such as on an incline, on a memory foam mattress, and through thick grass, and can be fine-tuned in each to further improve performance.
Our code to reproduce all results (including a real-world environment and training) is available on our website: \videourl 

\section{Related Work}
\label{sec:relatedwork}
Legged locomotion has long been of interest to roboticists, and a large body of work is dedicated to developing controllers by means of model-based optimal control~\cite{Kalakrishnan2010FastRQ, Bellicoso2018DynamicLT, dai2014whole, kuindersma2016optimization, Hutter2016ANYmalA, Bledt2018MITC3, Katz2019MiniCA}. Model-based methods have enabled a range of robotic locomotion skills, from high-speed running~\cite{Park2017BoundingCheetah} to backflipping~\cite{Ding2020RepresentationFreeMP} but require careful modeling of the conditions for which they will perform well in. Recent research has also achieved remarkable success by using learned approaches. One such approach that bypasses the complexities of real-world learning leverages simulation to train behaviors that transfer to the real world, demonstrating impressive results from robust walking~\cite{Tan2016SimulationbasedDO, Tan2018SimtoRealLA, He2018ZeroShotSC, Yu2019SimtoRealTF, Iscen2018PoliciesMT, Xie2019LearningLS, Hwangbo2019LearningAA, Lee2020LearningQL, Peng2020LearningAR, Yang2021FastAE, Rudin2021LearningTW, Kumar2021RMA} to more complex behaviors like agile running~\cite{margolisyang2022rapid}, jumping~\cite{Rudin2021CatLikeJA, Margolis2021LearningTJ} or bipedalism~\cite{Yu2022MultiModalLL, Vollenweider2022AdvancedST, Fuchioka2022OPTMimicIO, Smith2023LearningAA}. These approaches rely on zero-shot generalization, which has been shown to suffice in many scenarios of interest. However, they have two key limitations: they require extensive engineering of the simulation settings for the policy to generalize, and most do not have any mechanism to learn from their failures in the real world.

The alternative learning approach we study in this work is to directly learn in the environment of interest. Early work on learning \emph{directly} in the real world explored utilizing higher level actions in trajectory space~\cite{Kohl2004PolicyGR, Tedrake2004StochasticPG, Endo2005LearningCS, Luck2017FromTL, Choi2019TrajectorybasedPP, Yang2019DataER}, limiting their applicability to skills beyond walking or running at different speeds. Most similar to our setup are works that have studied real-world training using low-level PD target actions, with a 12-DoF A1 robot enabled by sample-efficient RL. Wu et al.~\cite{Wu2022DayDreamerWM} report learning to walk without resets and to recover from pushes in 1 hour using a model-based RL algorithm. We build our system on the simple model-free framework~\cite{Smith2022AWI} that demonstrated learning to walk from scratch in various environments, each within 20 minutes~\cite{Smith2022AWI}. While extremely efficient, the maximum achieved velocity in this prior work was less than 0.1 m/s, and the method did not demonstrate transfer between or adaptation to different terrains. Our work differs in three ways: (1) it leads to a \textbf{1.4x} improvement in the speed achieved after continued training; (2) it enables the resulting policy to perform better when directly transferred to more challenging situations; and (3) it demonstrates rapid adaptation through continued training when the policy does not generalize well in unseen settings.

A component of our approach is to regularize the actions produced by the policy according to model misprediction, which leads to fewer falls and lower torques during training. For locomotion, falls lead to physical damage and increased wall-clock training time. Ha et al.~\cite{Ha2020LearningTW} make this observation and thus use a constrained MDP (CMDP) formulation to explicitly combat falling during training by penalizing actions that lead to such states.
We find in~\autoref{sec:sim-experiments} that the CMDP approach still requires many failures in order to properly assign credit to which actions lead to fallen states. Our action regularization is similar insofar as it manifests as a penalty for the actor, but we directly regulate the magnitude of actions the policy explores within. Therefore, in~\metabbr, the policy does not need to first learn which actions lead to falls by taking them in the real world and experiencing failure. We leverage our knowledge that small actions around the nominal pose are unlikely to cause failure, and bias the policy to explore in this space first. 
While this strategy empirically leads to fewer falls than fully unbridled exploration, our primary goal is not to formally ensure safety but to enable efficient learning in the real world.

\section{System Design}
\label{sec:system}
We use the Markov Decision Process (MDP) formalism to define our task of learning to walk. An MDP is defined by a state space $\statespace \subset \mathbb{R}^n$, action space $\actionspace \subset \mathbb{R}^m$, initial state distribution $p_0(\cdot)$, transition function $p(\cdot|s, a)$, reward function $r: \statespace \times \actionspace \rightarrow \mathbb{R}$, and discount factor $\gamma \in [0, 1)$. RL maximizes the expected discounted cumulative return induced by the policy $\pi: \statespace \rightarrow \actionspace$:
\begin{align*}
  \expec_{\substack{a_t \sim \pi(\cdot| s_t)\\ s_{t+1} \sim p(\cdot| s_t, a_t)}} \left[ \sum_{t=0}^\infty \gamma^t r(s_t, a_t) \right].
\end{align*}
We use a type of algorithm that fits a critic $Q_{\theta}(\state, \action)$ to estimate this objective, and the policy then uses the critic to improve by maximizing the objective:
\begin{align}
  \mathcal{L}^{\tt SAC}_{\tt act}(\phi) = \expec_{\substack{s \sim D \\ a \sim \pi_\phi(\cdot|s)}} \Big[Q_{ \theta} (\state_{t},\action_{t}) \Big]. \notag
\end{align}

We build on the actor-critic RL method and implementation used by Smith et al.\cite{Smith2022AWI} (details in \textit{(b)}). We additionally use resets to improve plasticity~\cite{Nikishin2022ThePB, DOro2023SampleEfficientRL}. All neural networks are 2-layer feed-forward networks constructed and trained using JAX~\cite{jax2018github}. We use an Origin EON15-X laptop with an NVIDIA GeForce RTX 2070 GPU.

\paragraph{MDP formulation} We use the Go1 quadruped from Unitree for real-world experiments and perform analysis in simulation using the MuJoCo Menagerie model~\cite{menagerie2022github}. The robot's task is to walk as fast as possible without falling, where falling is defined as the robot's roll or pitch exceeding 30 degrees from upright. The robot's state $\state$ comprises its root orientation, joint angles and velocities, root (local) velocity, normalized foot contacts, and last recorded action. Previous works~\cite{Smith2022AWI, Wu2022DayDreamerWM} used a Kalman filter to fuse forward kinematics and acceleration to estimate the robot's velocity. In this work, we instead obtain velocity estimates from an Intel RealSense T265 camera attached to the robot's neck. We found the vision-based estimator to be more reliable and less susceptible to drift. We use target joint angles for the action space as is common for learned locomotion controllers on similar hardware~\cite{Peng2020LearningAR, Kumar2021RMA, Fu2021MinimizingEC}. The control frequency is 20Hz. The target joint angles are fed to a PD controller, with a position gain of 20 and derivative gain of 1, which converts them to torques at the leg joints. We also employ two forms of smoothing: a low-pass filter on the policy outputs and action interpolation, which is running at 500Hz. We define a reward function to maximize the robot's local linear velocity while maintaining an upright orientation. We also include penalties on angular velocity and torque smoothness. The exact definition of all reward terms can be found on our website.

\label{sec:method}

\begin{algorithm}[t]
    \caption{\texttt{\metabbr} pseudocode}
    \label{alg:method}
    \begin{algorithmic}[1]{
    \footnotesize
        \Require Action regularization configs: growth rate $N_{\text{curr}}$, initial range $\mathcal{A}_{\text{start}}$, end range $\mathcal{A}_{\text{end}}$, penalty weight $\sigma$, dynamics shift threshold $\Delta_M$, replay ratio $rr$, max gradient steps $\nabla_M$.
        \State Initialize parameters of $Q_\theta$ and $\policy_\phi$ and a replay buffer $\mathcal{B} \leftarrow \emptyset$
        \State Initialize action regularization states $i = 0, \mathcal{A}_{\text{i}} = \mathcal{A}_{\text{curr\_i}}, \mathcal{A}_{\text{e}} = \mathcal{A}_{\text{curr\_e}}$
  	\Repeat
            \State Collect transition $(\state_{t}, \action, \state_{t+1}, r_t)$ with $\policy_\phi$ and store in $\mathcal{B}$
            \item[]
            \State \textsc{// Update regularization}
            \State Increment counter $i+=1$
            \State Determine progress $\alpha_{\text{curr}} = \text{clip}(i/N_{\text{curr}}, 0, 1)$
            \State Calculate corresponding space $\mathcal{A}_{\text{curr}} \leftarrow \alpha_{\text{curr}}\mathcal{A}_{\text{e}} + (1-\alpha_{\text{curr}}) \mathcal{A}_{\text{i}}$
            \State Calculate dynamics error $\Delta_{\text{curr}} \leftarrow \big(\state_{t+1} - \hat{f}_{\psi}(\state_{t}, \action_{t})\big)^2$
            \If{ $\Delta_{\text{curr}} \geq \Delta_M$ }
                \State Set $i \leftarrow 0$, $\mathcal{A}_{\text{i}} \leftarrow 0.9 \times \mathcal{A}_{\text{curr}}$
            \EndIf
            \item[]
            \State \textsc{// Perform updates}
            \For{$rr$ steps}
                \State Update $\theta$ with critic loss
            \EndFor
            \State Update $\psi$ with $\mathcal{L}^{\tt dyn}(\psi)$ in ~\autoref{eq:dyn-mse}
            \State Update $\phi$ with $\mathcal{L}^{\tt APRL}_{\tt act}(\phi)$ in~\autoref{eq:aprl_actor}
            \item[]
            \State \textsc{// Periodically reset weights}
            \If{ $i \cdot rr > \nabla_M$}
                \State Reinitialize $\theta, \phi, \psi$ and reset $i = 0$
            \EndIf
  	\Until{forever}
    }\end{algorithmic}
\end{algorithm}

\begin{figure}[t]
    \centering
    \includegraphics[width=\linewidth]{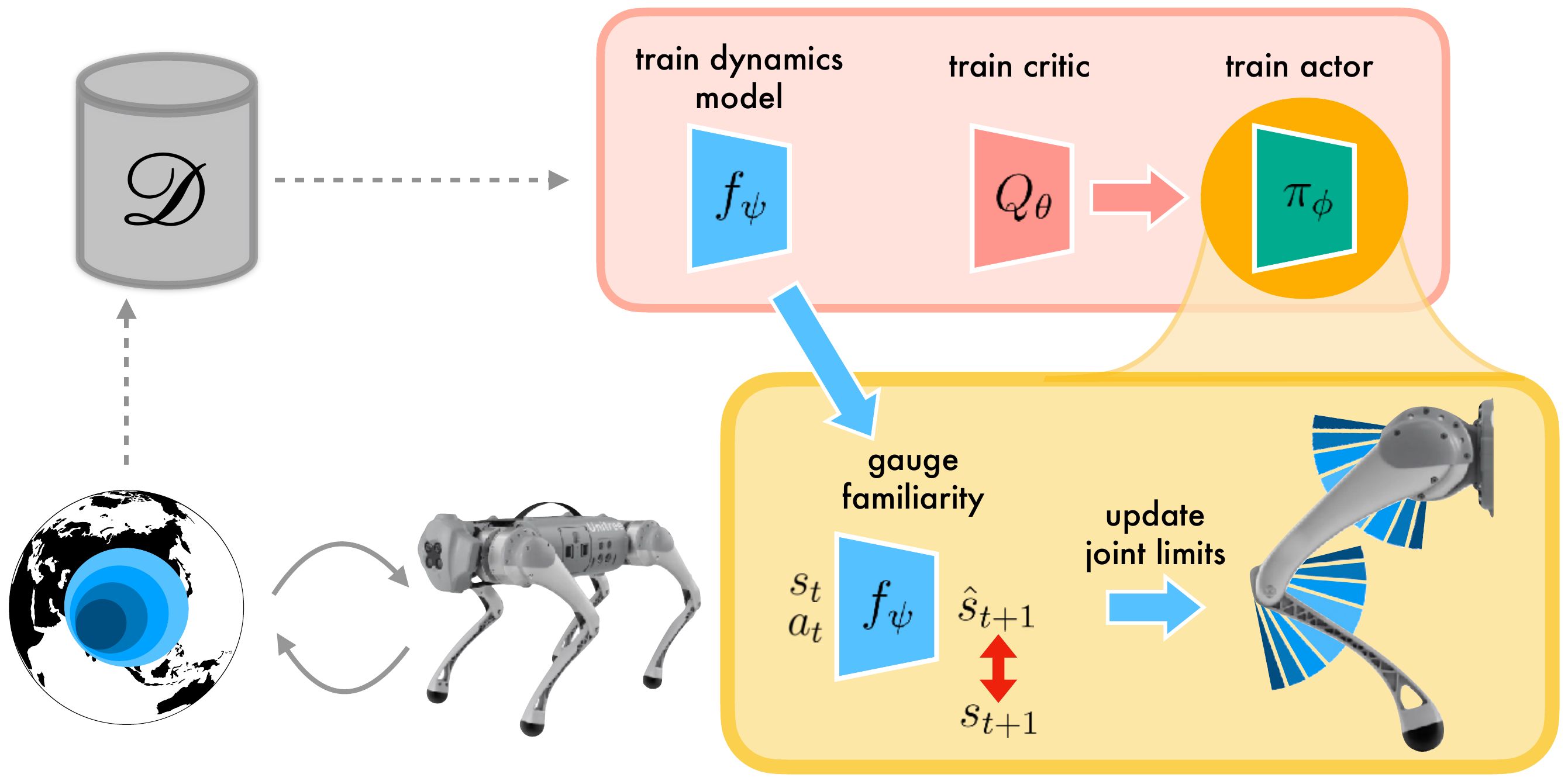}
    \caption{\textbf{Overview of \metabbr.} We train the robot with a flexible constraint, represented by blue semicircles around the joints. It collects experience, storing it in a replay buffer for training an actor and critic, as explained in~\autoref{sec:system}, alongside a predictive dynamics model. The model's prediction error adjusts the constraint's bounds, either tightening (for high error) or relaxing (for accurate predictions). This adjustment is incorporated into the actor's loss, as specified in~\autoref{eq:aprl_actor}.}
    \label{fig:overview}
    \vspace{-1mm}
\end{figure}

\paragraph{Replay ratio and resets} Sample-efficient RL methods use a high \textit{replay ratio}, i.e., the ratio of gradient updates to real-world transitions collected, to make efficient use of the collected data. To be able to take more updates on the same data, they require some form of regularization, e.g., using model-generated data~\cite{Janner2019WhenTT, Hafner2020DreamTC}, weight-based regularization~\cite{Liu2020RegularizationMI, Hiraoka2021DropoutQF}, ensembling at the agent level~\cite{Lee2020SUNRISEAS} or at the critic level~\cite{Chen2021RandomizedED}, or a combination of techniques~\cite{Li2023EfficientDR}. In this work, we use a high replay ratio with a model-free method, using Dropout~\cite{Srivastava2014DropoutAS} and LayerNorm~\cite{Ba2016LayerN} to regularize the critic. Nikishin et al.~\cite{Nikishin2022ThePB} showed that excessive training on early data with high replay ratio methods can cause the networks to lose plasticity, the ability to continue learning with more data, and propose periodic resets of the agent to mitigate this effect. Resetting specifically implies reinitialization of network weights and optimizer states while maintaining the replay buffer. We incorporate this regularizer into our adaptive strategy as we will describe next.

\section{Efficient Learning of Legged Locomotion with Adaptive Policy Regularization}
We present our system for efficiently learning and fine-tuning quadrupedal locomotion in real-world scenarios using \textbf{A}daptive \textbf{P}olicy \textbf{R}egu\textbf{L}arization (APRL). Our framework, shown in Figure~\ref{fig:overview}, involves dynamically modulating policy regularization over the course of training to provide the policy with adequate room to explore and improve, but not so unbridled as to lead to inefficient---and often violent---training. To do so, we introduce `soft' constraints on the actions (defined in \textit{(b)}) that are adjusted based on how `familiar' the robot is in its current situation (described in \textit{(c)}). We also incorporate resets to improve plasticity, i.e., the ability to keep learning from new data. In the remainder of this section, we describe the principle underlying our choice of regularization. We then detail how we adapt the constraints based on the robot's learning progress and finally how we implement them in practice. \autoref{alg:method} summarizes the training procedure in pseudocode.

\paragraph{An efficiency-performance trade-off}
Prior work has shown that explicit action limits have an enormous effect on learning speed in the real world~\cite{Smith2022AWI}---intuitively, limiting the policy's search space makes finding a solution faster. In the case of legged locomotion, where policies are often parameterized to output PD targets as actions~\cite{Peng2020LearningAR, Kumar2021RMA, Fu2021MinimizingEC, Smith2023LearningAA}, searching in a limited region around a nominal pose is reasonable since the policy may still learn to walk at a low speed and is less prone to falling. These details are important for real-world learning, as each fall incurs nontrivial physical damage and time costs (specifically, an additional 5-10 seconds which translates to an opportunity cost of 100-200 time steps). Beyond falling, large changes in PD targets translate to large torques, which can cause significant damage to the robot over time. As we show in our experiments, while a restricted action space is helpful for learning to walk quickly, it can significantly inhibit the learned policy's ultimate capabilities.

\paragraph{Soft policy constraints} \label{sec:define-action-space} A straightforward approach to constraining the policy's actions is to use a transformation that either maps any action beyond the limits to the corresponding extremum or scales the actions within fixed bounds. We find that these na\"ive implementations do not work well in practice (see~\autoref{sec:sim-experiments}) for our application. We instead introduce an action penalty, which can be interpreted as a soft constraint on the policy. Specifically, each action dimension is a target joint angle, where the minimum and maximum correspond to the lower and upper physical limits, respectively. We define a feasible region $~\mathcal{A}_\epsilon$ that corresponds to actions whose elementwise magnitude is no greater than $\epsilon$, which can write as an inequality constraint on each dimension $i$: $
c_i(\action, \epsilon) = |\action_i| - \epsilon \leq 0 $.
Using a penalty method amounts to training with a penalty on infeasible iterates, and we found an L1 penalty with fixed weight $\sigma=10$ to work best in our case. This leads to the modified actor loss:
\begin{align}
  \mathcal{L}^{\tt APRL}_{\tt act}(\phi) = \expec_{\substack{s \sim D \\ a \sim \pi_\phi(\cdot|s)}} \Big[Q_{ \theta} (\state_{t},\action_{t}) - \textcolor{myblue}{\sigma \sum_i \max(0, c_i(\action_t))} \Big]. \label{eq:aprl_actor}
\end{align}

\paragraph{Deciding on appropriate feasible regions} As previously mentioned, we would intuitively like to allow the robot to be more exploratory when in a familiar setting. Conversely, we would like to constrain the robot to promote more conservative behavior when the robot is in a setting that is different from that in which it was trained. To do this, we use a dynamics model: we fit $\hat{p}_\psi(\state'\mid\state, \action)$ on the data the robot collects and use the likelihood of the latest observed transitions to determine whether a situation is `familiar'. We specifically choose this heuristic over others, e.g., model disagreement to measure epistemic uncertainty~\cite{Shyam2018ModelBasedAE}, to implicitly encourage the policy to favor predictable solutions. Dynamics prediction error also explicitly detects changes in the environment dynamics, which we would like the robot to be able to immediately react and adapt to. We represent the dynamics model as $\hat{p}_\psi(\state'\mid\state, \action)=\mathcal{N}(\hat{f}_\psi(\state, \action), I)$, where $f_\psi$ is a neural network. Training with maximum likelihood corresponds to a mean squared error (MSE) loss:
\begin{align}
  \mathcal{L}^{\tt dyn}(\psi) = \expec_{(\state,\action,\state') \sim D} \Big[(\hat{f}_{\psi}(\state,\action) - \state')^2 \Big]. \label{eq:dyn-mse}
\end{align}
We use a schedule such that the joint limits are grown at every time step by a constant increment (line 8) unless the likelihood of the most recent data is registered as highly unlikely by the learned dynamics model, i.e., if the MSE incurred by the dynamics model surpasses a maximum of $\Delta_M$ (lines 9-10). If this threshold is hit, we shrink the joint limits by a multiplicative factor (line 11) to allow the robot to react quickly in a new situation. The exact hyperparameters used in our experiments can be found on our website.

\section{Real-World Results}
\label{sec:real-world-experiments}
Our real-world experiments test whether~\metabbr can enable a real quadrupedal robot to efficiently learn to walk entirely in the real world and adapt to new dynamics that are more challenging than demonstrated by prior work. We specifically seek to answer the following questions: 
\begin{enumerate}
    \item Can we enable a real 12 DoF quadrupedal robot to learn to walk in a matter of minutes \emph{without} a manually defined, Restricted action space?
    \item Does~\metabbr enable \textit{continued improvement} as the robot collects more data?
    \item Does~\metabbr enable the robot to learn to walk in more challenging settings?
    \item Can we use~\metabbr to allow the robot to continue learning amid changing dynamics?
\end{enumerate}
\subsection{Experimental Setup}
\begin{figure}[t]
    \centering
    \includegraphics[width=1\linewidth]{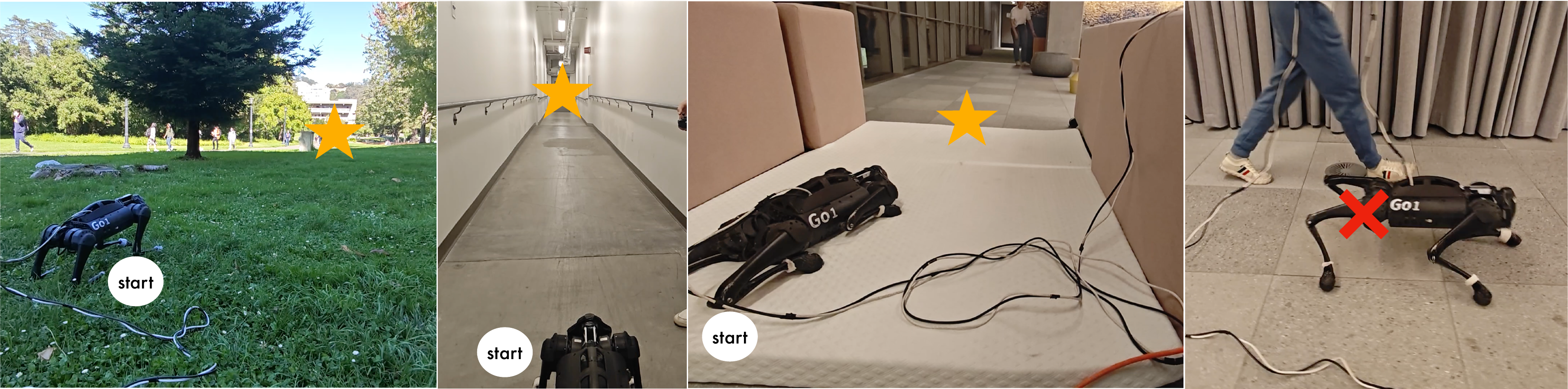}
    \caption{\footnotesize \textbf{Environment visualization.} We visualize the different environments we test in: Grass, Ramp, Mattress, and Frozen Joint. We indicate the start and goal locations for the situations in which there is a path for the robot to traverse and is evaluated on its time to finish the entire path.}
    \label{fig:real-task-visualization}
\vspace{-0.5em}
\end{figure}

We address the first two questions by comparing to the prior work of Smith et al.~\cite{Smith2022AWI} (labeled as \textit{Restricted [46]}), as this prior method also focuses on learning to walk directly through real-world training, in the same, flat-ground environment training for 80k steps each (roughly 80 minutes of real-world interaction time). 
To address (3) and (4), we evaluate the learned policies in 4 new scenarios (shown in~\autoref{fig:real-task-visualization}):
\begin{itemize}
    \item \textbf{Mattress}: The robot must walk across a 5cm thick memory foam mattress. The robot's feet sink and the depression of the mattress makes walking more difficult, requiring a unique gait with more force.
    \item \textbf{Ramp}: We task the robot to walk up a slippery, 5-degree inclined ramp. The inclination and slipperiness require the robot to maintain good balance while giving strong pushes on the back legs to climb up.
    \item \textbf{Grass}: We deploy the robot outdoors on grass. The unevenness of the mud underneath and the unique friction properties require good foot clearance and quick response to changes in the shape of the terrain.
    \item \textbf{Frozen Joint}: We freeze the thigh joint on the rear right leg to test adaptation to sudden shifts in dynamics in a controlled way, where there are no natural variations inn terrain to cause differences in performance.
\end{itemize}
In these evaluations, to account for stochasticity and variance in the real world, we run each evaluation at least 3 times and report the mean and standard deviation across these runs.
For (3), we first test the policies without any fine-tuning. We introduce two additional evaluation metrics: the time each policy requires to traverse from one end of a path to another (pictured in \autoref{fig:real-task-visualization} as the white circle and gold star, respectively) and the number of times the robot fell while doing so. For these evaluations, we manually reset the robot onto the path if it veers too off-course. Note though that we do not include the relative finish time and fall counts for the ``Frozen Joint" scenario because the shift in dynamics is not in terrain.
Lastly for (4), we evaluate whether a few minutes of continued training (specifically 3k time steps) allows the robot to improve in these scenarios. That is, we fine-tune in the target setting, then re-evaluate using the same protocol.

\begin{figure}[t]
    \centering
    \includegraphics[width=\linewidth]{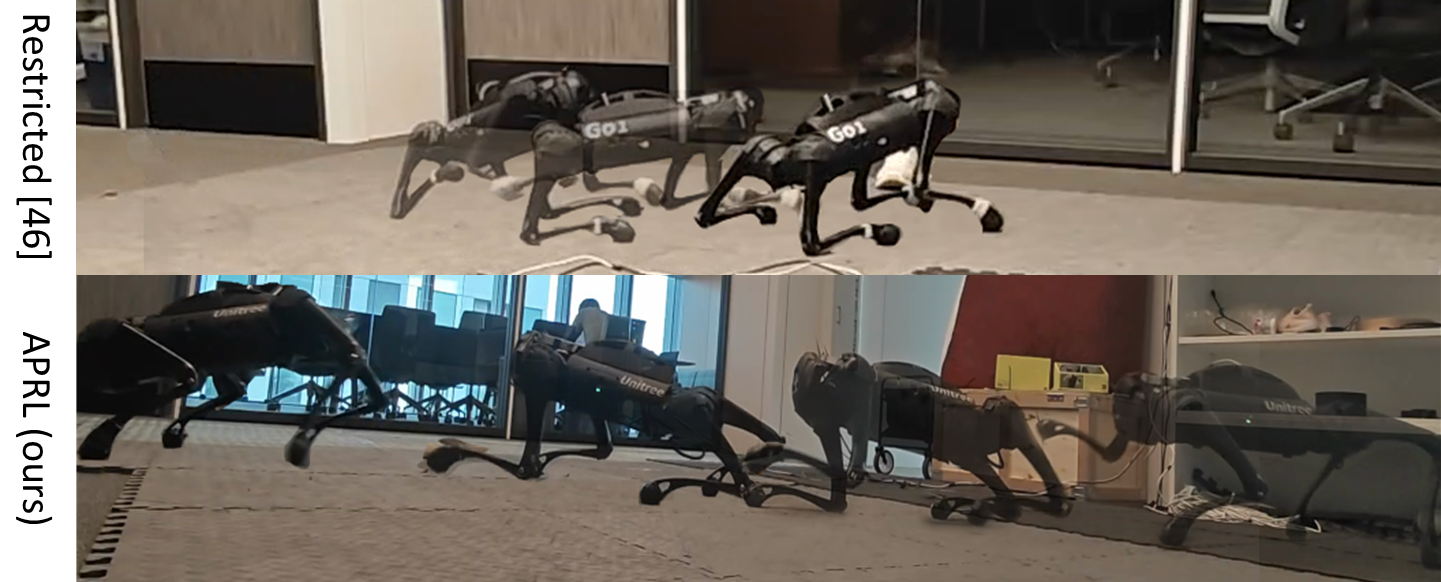}
    \caption{\footnotesize \textbf{Qualitative comparison of policies.} We compare the gaits learned, (top) Restricted and (bottom) \metabbr, from scratch on flat ground by showing a time-lapse of the policies rolled out for 5 seconds each. Our policy learned to use its front legs to step and propel its back legs in a cantering-like manner whereas the Restricted policy drags and slides across the ground.}
    \label{fig:real-training-qualitative}
\end{figure}
\begin{figure}[t]
    \centering
    \includegraphics[width=1\linewidth]{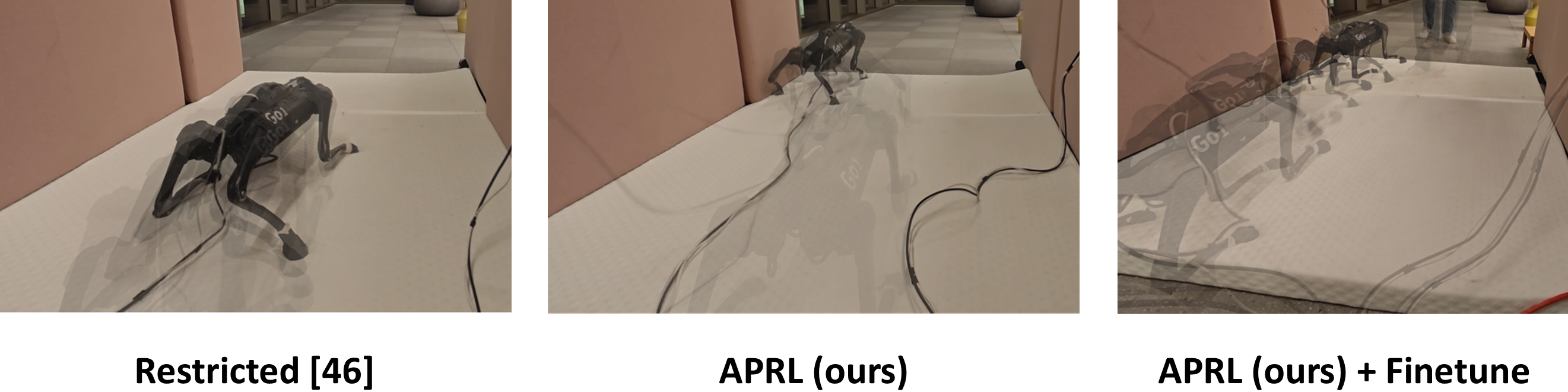}
    \caption{\textbf{Qualitative comparison on new terrain.} We show a 5 second time-lapse of evaluating different policies on the mattress. The Restricted method tries to slide on the mattress, which slows it down significantly. \metabbr policies have a higher foot clearance, so they are able to traverse it more efficiently and, after fine-tuning, with fewer falls.}
    \label{fig:real-matress-timelapse}
\end{figure}

\subsection{Results}

\begin{wrapfigure}[15]{R}{0.2\textwidth}
    \centering
    \vspace{-10mm}
    \includegraphics[width=0.2\textwidth]{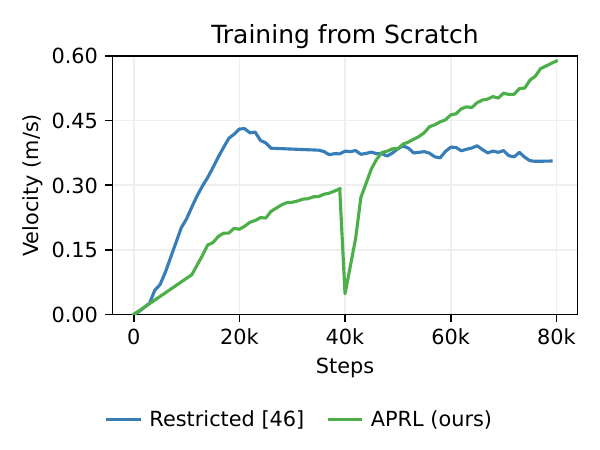}
    \caption{\textbf{Forward velocity achieved during training.} The Restricted method learns more efficiently at first but is unable to keep improving. Meanwhile, our method still learns to walk quickly but acquires a maximum velocity of 0.62 m/s. Note that the dip at 40k steps is due to parameter resets to improve plasticity (line 19 in~\autoref{alg:method}).}
    \label{fig:real-training-quantitative}
\end{wrapfigure}

\noindent\textbf{Training from scratch.} In \autoref{fig:real-training-quantitative}), we see that even without manual restriction of the action space, \metabbr starts learning to walk immediately. We attempted to compare to training without regularization; however, the robot's actions were too aggressive to collect even a small amount of data. \metabbr's adaptive regularization makes training possible in way that, importantly, allows the robot to \emph{continue to improve}, achieving a maximum average velocity of 0.62 m/s. In contrast, the Restricted method indeed learns to walk extremely quickly but plateaus early in training and achieves a maximum average velocity of only 0.44 m/s. This performance almost matches that of its simulated variant (see~\autoref{sec:sim-experiments}), so we believe this cap to be a fundamental limitation rather than a real-world-specific challenge. \metabbr's performance also closely tracks its simulated variant's; however, we hypothesize that our performance is limited in the real world partially due to space constraints, as the robot is only able to take a few steps before reaching the workspace limits once it reaches a higher speed. We also observe that \metabbr produces a visually more naturalistic gait, shown in \autoref{fig:real-training-qualitative} and better viewed in video form on our project website. These results show that \metabbr is significantly better equipped than na\"ive RL to continually improve as it collects more data, as opposed to quickly reaching but plateauing with limited capabilities.

\begin{figure}[t]
    \centering
    \includegraphics[width=0.7\linewidth]{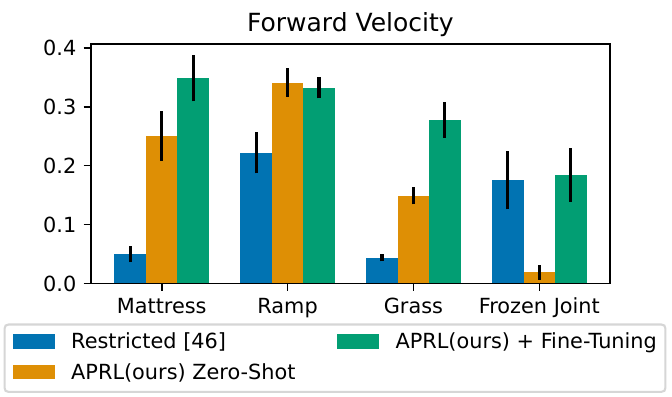}
    \caption{\footnotesize \textbf{Real-world velocity comparisons (higher is better).} In all scenarios except frozen joint, \metabbr significantly outperforms the Restricted method in terms of its velocity when tested in new scenarios. With just minutes of fine-tuning, \metabbr significantly improves performance in all settings except on the ramp, where it is comparable.}
    \label{fig:real-eval-velocities}
    \vspace*{-4mm}
\end{figure}

\begin{figure}[t]
    \centering
    \includegraphics[width=1\linewidth]{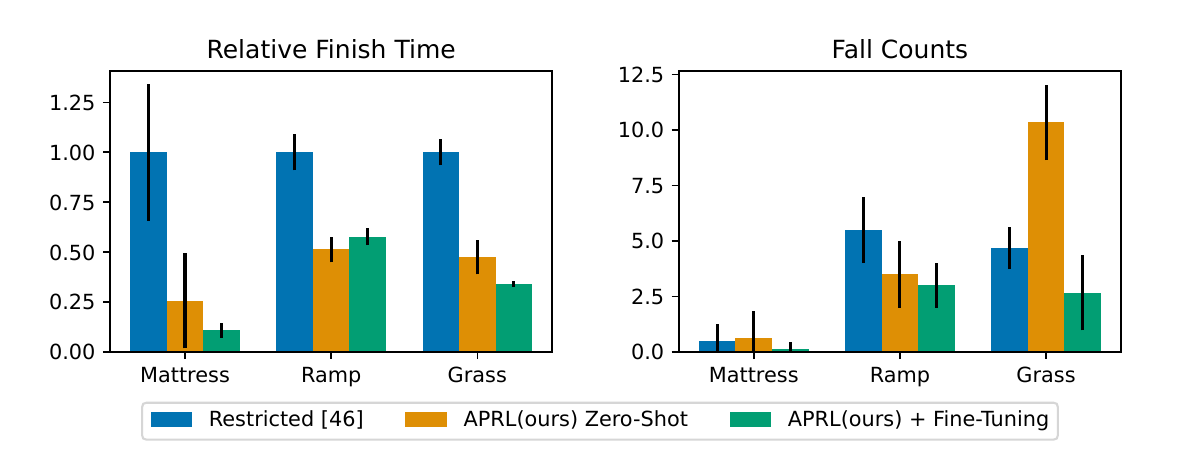}
    \caption{\footnotesize \textbf{Real-world finish time and fall count (lower is better).} (Left) We report the time taken to traverse a path relative to the Restricted method and absolute fall count. On Ramp and Grass, \metabbr is \textbf{2x} faster, and on Mattress, \metabbr is \textbf{almost 4x} faster.}
    \label{fig:real-eval-other-metrics}
\end{figure}

\noindent\textbf{Transferring to different scenarios.} 
We find that \metabbr not only successfully enables a quadrupedal robot trained only in the real world to walk amid a variety of conditions, but also to \emph{keep improving} as it continues to be deployed. Quantitatively, the policy learned with \metabbr even without fine-tuning is significantly better on average at walking than the Restricted policy in terms of average velocity (see \autoref{fig:real-eval-velocities}) and at completing a given path faster and with fewer falls (see \autoref{fig:real-eval-other-metrics}). The exception is when we freeze a joint, in which case the Restricted policy generalizes much better during zero-shot evaluation. In this case, we find that with continued training, \metabbr can quickly learn to overcome this gap. In \autoref{fig:real-matress-timelapse}, we show a qualitative comparison of policies where the path can be visualized with a static camera. We encourage the reader to view the qualitative differences in policies for each scenario on our project website. 

\section{Simulated Analysis}
\label{sec:sim-experiments}
In this section, we analyze~\metabbr using a simulated version of the task described in~\autoref{sec:system}. Although simulation does not model many of the real-world complexities that we aim to address, we use it to perform controlled experiments for comparison purposes and insight. We design our simulated experiments to answer the following questions:
\begin{enumerate}
    \item Does restricting the action space actually cap the robot's achievable velocity?
    \item If so, does~\metabbr allow overcome these limits, and how does it compare to `optimal' behavior? 
    \item How does~\metabbr compare to prior work and ablations?
\end{enumerate}

\begin{figure}
    \centering
    \includegraphics[width=1\linewidth]{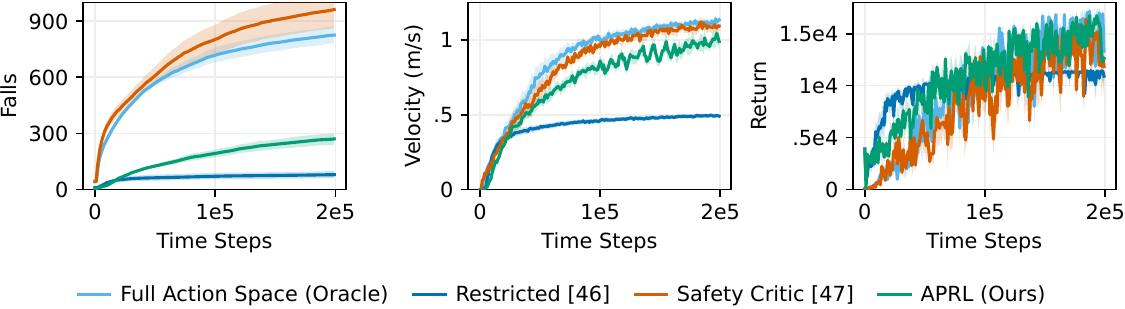}
    \caption{\footnotesize \textbf{Comparisons.} We report each policy's performance measured by the falls, average velocity, and return (mean and standard error across 5 seeds) with respect to the number of time steps. We find that \metabbr is the only method that effectively balances achieving high velocity while regulating the number of falls such that it is feasible to run in the real world. }
    \label{fig:sim-actspace-comparison}
\end{figure}

To answer (1), we compare learning with a fixed, limited action range as done in the Walk in the Park system~\cite{Smith2022AWI} (labeled \textit{Restricted}) to learning with the full action range without our adaptive regularization (labeled \textit{No Regularization}). As mentioned in \autoref{sec:real-world-experiments}, this comparison is not feasible in the real world but gives the policy the most freedom to optimize, so we use its asymptotic performance as the upper bound on the robot's capabilities. We see in ~\autoref{fig:sim-actspace-comparison} that the way actions are explored has a profound effect on training performance. Restricted excels at the very start of training, achieving a steeper learning curve and fewer falls than No Regularization, but also saturates very quickly and is not capable of improving beyond a velocity of about 0.5 m/s. In contrast, the policies with access to the full joint range eventually far exceed the hard-constrained policy's performance as measured by return. 

For (2), we observe that~\metabbr achieves near-optimal performance in comparing its asymptotic performance in terms of return and achieved velocity, with that of the non-regularized `oracle'. Furthermore, it does so with significantly fewer falls, making it feasible to run in the real world. To answer (3), we include a comparison to the constrained MDP formulation of Ha et al.~\cite{Ha2020LearningTW} by training a critic to predict falls and penalizing the policy for taking actions that the critic predicts will lead to falls. We found that in our case the safety critic required tens of thousands of samples to converge, which was not quick enough to shape the early stage of exploration to prevent excessive falls. In fact, this method was especially sensitive to network initialization and so we omit one seed that diverged for clarity. Generally, it is quite difficult to fit a critic with time-delayed effects whereas our method simply penalizes action magnitudes directly. This method was shown to be effective with a significantly simpler robot, with removed degrees of freedom, where learning a critic may be expected to be much simpler than in our setting.

Finally, we compare~\metabbr's adaptive action regularization to alternatives in order to understand the importance of (a) using a soft constraint rather than a na\"ive hard constraint (b) using an adaptive rather than fixed expansion rate and (c) regularizing the policy directly rather than through the reward function. For (a), we compare to \textit{Hard Constraint}, which is our method but instead of penalizing the policy outputs, we clip them at the prescribed limits before applying them in the environment. Next, for (b) we test whether the prediction error is meaningfully regulating the speed at which the constraint is changing by only using the constant increment (removing line 11 of~\autoref{alg:method}) and label this as \textit{Non-Adaptive Regularization}. Lastly, for (c) we implement a baseline in which we add a very common control cost to the reward function and call this \textit{Reward Regularization}---we follow standard conventions and use a quadratic penalty on the actions. From Fig. \ref{fig:sim-actspace-comparison}, we see that with non-adaptive regularization there are too many falls which is a major issue for real-world deployment. This problem is exacerbated even further when using a hard constraint. Adding the action penalty via reward regularization causes the policy to exploit the reward getting high return, but with no forward velocity so it does not actually perform the task.
\begin{figure}
    \centering
    \includegraphics[width=1\linewidth]{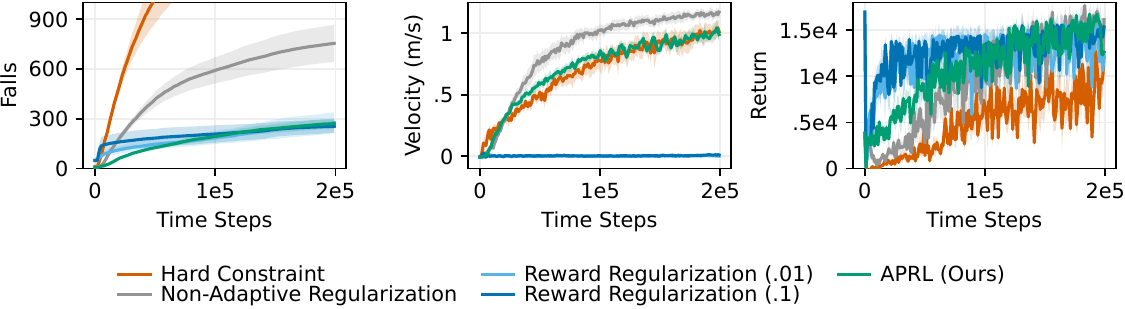}
    \caption{\footnotesize \textbf{Ablations.} We compare to versions of \metabbr that use (a) a hard constraint (b) non-adaptive regularization (c) regularization via the reward function. These either have too many falls or do not progress on \textit{forward velocity}, showing the importance of all design components of \metabbr.}
    \label{fig:sim-actspace-differentways}
\end{figure}

\section{Conclusion}
\label{sec:conclusion}
We presented a system for efficiently learning quadrupedal locomotion directly in the real world that improves upon prior work in terms of efficiency and final achieved performance. \metabbr introduces adaptive policy regularization that encourages the policy to explore within action limits that are commensurate to the policy's competence in a particular situation. \metabbr allows the robot to effectively utilize its full joint range without causing excessive falling during training. The final speed attained by our policies improves significantly over prior work, both when training from scratch and when fine-tuning to a new terrain. 

Our method has several limitations. Although our regularization technique reduces the number of falls, it does not provide a proper ``safety'' mechanism in the sense that it does not aim to prevent all failures. While this is not a major issue for the small quadrupedal robot we use, it might be a more severe challenge for larger robots. Our method also does not utilize any visual perception, and incorporating this might present additional challenges for sample complexity. Lastly, although the final speed and gait quality acquired by our method improves significantly over the prior approach that learns directly in the real world, the quality of the gaits is still lower than those that have been demonstrated in simulation. Addressing these limitations is an important direction for future work, but we hope that our demonstrated results already indicate that our method represents an important step toward robotic systems that can continually adapt in the real-world at deployment time, such that we do not need to train policies that never fail, but can instead allow them to learn to avoid mistakes after they happen.

\subsection*{Acknowledgements}
This work was supported in part by ARO W911NF-21-1-0097, ARL DCIST CRA W911NF-17-2-0181, and ONR N00014-20-1-2383 and N00014-22-1-2773.
Laura Smith is supported by the Google PhD Fellowship. 
We thank Russell Mendonca, Philipp Wu, Kevin Zakka, Ademi Adeniji, Oier Mees, Seohong Park, Zhongyu Li, and Qiyang Li for their valuable feedback. We thank Philipp Wu for designing and printing \href{https://github.com/wuphilipp/robot_parts}{the protective shell} and Kevin Zakka for providing support with the simulated robot model.
\balance
\bibliography{references}
\bibliographystyle{IEEEtran}

\end{document}